\documentclass[journal]{IEEEtran}
\ifCLASSINFOpdf
   \usepackage[pdftex]{graphicx}
   \graphicspath{{../pdf/}{../jpeg/}}
   \DeclareGraphicsExtensions{,.jpeg,.png}
\else
   \usepackage[dvips]{graphicx}
   \graphicspath{{../eps/}}
   \DeclareGraphicsExtensions{.eps}
\fi

\usepackage{times}  
\usepackage{helvet}  
\usepackage{courier}  
\usepackage{url}  
\usepackage{graphicx}  

\usepackage{pdfpages}
\usepackage{mathrsfs}
\usepackage{amsfonts}
\usepackage{amsmath}
\usepackage{tikz}
\usepackage{amssymb}
\usepackage{textcomp}
\usepackage{subfigure}
\usepackage{verbatim}

\usepackage{multirow}
\usepackage[switch]{lineno}
\usepackage{fancyhdr}

\begin{document}
%
\title{Attributes Guided Feature Learning for\\ Vehicle Re-identification}

\author{Hongchao Li,~\IEEEmembership{}
        Xianmin Lin,~\IEEEmembership{}
        Aihua Zheng,~\IEEEmembership{}
        Chenglong Li,~\IEEEmembership{}
        Bin Luo*,~\IEEEmembership{}
        Ran He,~\IEEEmembership{}
        and~Amir Hussain~\IEEEmembership{}

\thanks{H. Li, X. Lin, and B. Luo are with School of Computer Science and Technology, Anhui University, Hefei 230601, China.}
\thanks{A. Zheng, and C. Li are with Information Materials and Intelligent Sensing Laboratory of Anhui Province, Anhui Provincial Key Laboratory of Multimodal Cognitive Computation, School of Artificial Intelligence, Anhui University, Hefei, 230601, China.}
\thanks{R. He is with Institute of Automation, Chinese Academy of Sciences, Beijing 100190, China.}
\thanks{Amir Hussain is with Edinburgh Napier University, School of Computing, Edinburgh EH10 5DT, Scotland, U.K.  }
\thanks{This research is supported in part by the National Natural Science Foundation of China (No. 61976002, 61976003, 62076003 and 61860206004), and the Joint Funds of the National Natural Science Foundation of China (No.U20B2068).}
\thanks{ * Corresponding author.}
}

\markboth{Journal of \LaTeX\ Class Files}%
{Shell \MakeLowercase{\textit{et al.}}: Bare Demo of IEEEtran.cls for IEEE Journals}

\maketitle
\thispagestyle{fancy}
\fancyhead{}
\lhead{}
\lfoot{}
\cfoot{\footnotesize \copyright~2021~IEEE. Personal use is permitted, but republication/redistribution requires IEEE permission.\\See http://www.ieee.org/publications standards/publications/rights/index.html for more information.}
\rfoot{}

\begin{abstract}
Vehicle Re-ID has recently attracted enthusiastic attention due to its potential applications in smart city and urban surveillance. However, it suffers from large intra-class variation caused by view variations and illumination changes, and inter-class similarity especially for different identities with a similar appearance.
To handle these issues, in this paper, we propose a novel deep network architecture, which guided by  meaningful attributes including camera views, vehicle types and colors for vehicle Re-ID. In particular, our network is end-to-end trained and contains three subnetworks of deep features embedded by the corresponding attributes.
For network training, we annotate the view labels on the VeRi-776 dataset. Note that one can directly adopt the pre-trained view (as well as type and color) subnetwork on the other datasets with only ID information, which demonstrates the generalization of our model.
Extensive experiments on the benchmark datasets VeRi-776 and VehicleID suggest that the proposed approach achieves the promising performance and yields to a new state-of-the-art for vehicle Re-ID.
\end{abstract}

\begin{IEEEkeywords}
Vehicle Re-identification, Deep Features, Attributes.
\end{IEEEkeywords}

%
\IEEEpeerreviewmaketitle

\section{Introduction}
%
%
%
%
\IEEEPARstart{V}{ehicle} re-identification (Re-ID) is a frontier and important research problem in computer vision, which has many potential applications, such as intelligent transportation, urban surveillance and security since vehicle is the most important object in urban surveillance.
The aim of vehicle Re-ID is to identify the same vehicle across non-overlapping cameras. Although license plate can uniquely identify the vehicle, it is scarcely recognizable due to the challenging factors of motion blur, challenging camera view and low resolution, to name a few.
Some researchers have explored spatio-temporal information~\cite{shen2017learning,liu2016deep,liu2018provid,hsu2019multi} to boost the performance of appearance based vehicle Re-ID.
However, it is difficult to obtain the complete spatio-temporal information since the vehicles may only appear in a few of the cameras in the large scale camera networks. Therefore, the prevalent vehicle Re-ID methods still focus on appearance based models.

Extensive works dedicate on person Re-ID in the past decade~\cite{zheng2017unlabeled,liu2018pose,wei2018person,chen2018deep,su2015multi,khamis2014joint,fan2018unsupervised}, which focus on two mainstreams:
(1) Appearance modelling~\cite{liu2018pose,wei2018person,fan2018unsupervised}, which develops robust feature descriptors to encode the various changes and occlusions among different camera views.
(2) Learning-based methods~\cite{zheng2017unlabeled,ahmed2015improved,liao2015person,chen2018deep,su2015multi,khamis2014joint}, which learns metric distance to mitigate the appearance gaps between the low-level features and high-level semantics.
Recently, deep neural networks have made a marvelous progress on both feature learning~\cite{zheng2017unlabeled,su2015multi,khamis2014joint} and metric learning~\cite{ahmed2015improved,liao2015person,chen2018deep} for person Re-ID.
However, directly employing person Re-ID models for vehicle Re-ID could not guarantee the satisfactory performance, since the appearance of pedestrians and vehicles varies in the different manner from different viewpoints.

Although much progress has been made on vehicle Re-ID~\cite{liu2016large,li2017deep,liu2016deep,liu2018provid,zhang2017improving,shen2017learning,tang2019cityflow,lou2019veri,he2019part,tang2019pamtri},
it still encounters many challenges as in addition to the common challenges in person Re-ID such as occlusion and illumination, etc.
The first crucial challenge of vehicle Re-ID is the large intra-class variation caused by the viewpoint variation across different cameras, which has been widely explored in person Re-ID~\cite{saquib2018pose,zhao2017spindle,zhao2017deeply,su2017pose}. This issue is even more challenging in vehicle Re-ID since most of the vehicle images under a certain camera are almost in the same viewpoint due to the rigid motion of the vehicles.
Unfortunately, it might not achieve the satisfactory performance when directly employing the methods from person Re-ID to vehicle Re-ID since the appearance distributes totally different between persons and vehicles.
Some vehicle Re-ID methods~\cite{zhou2018cross,zhou2018aware} use adversarial learning schemes to generate multi-view images or features from a single image, and can thus address the challenge of view variation to some extent.
But they might be difficult to distinguish different vehicles with very similar appearance. Furthermore, they neglect the attributes information, such as type and color, which would be critical cues for boosting the performance of vehicle Re-ID.

\begin{figure*}[t]
\includegraphics[width =18cm]{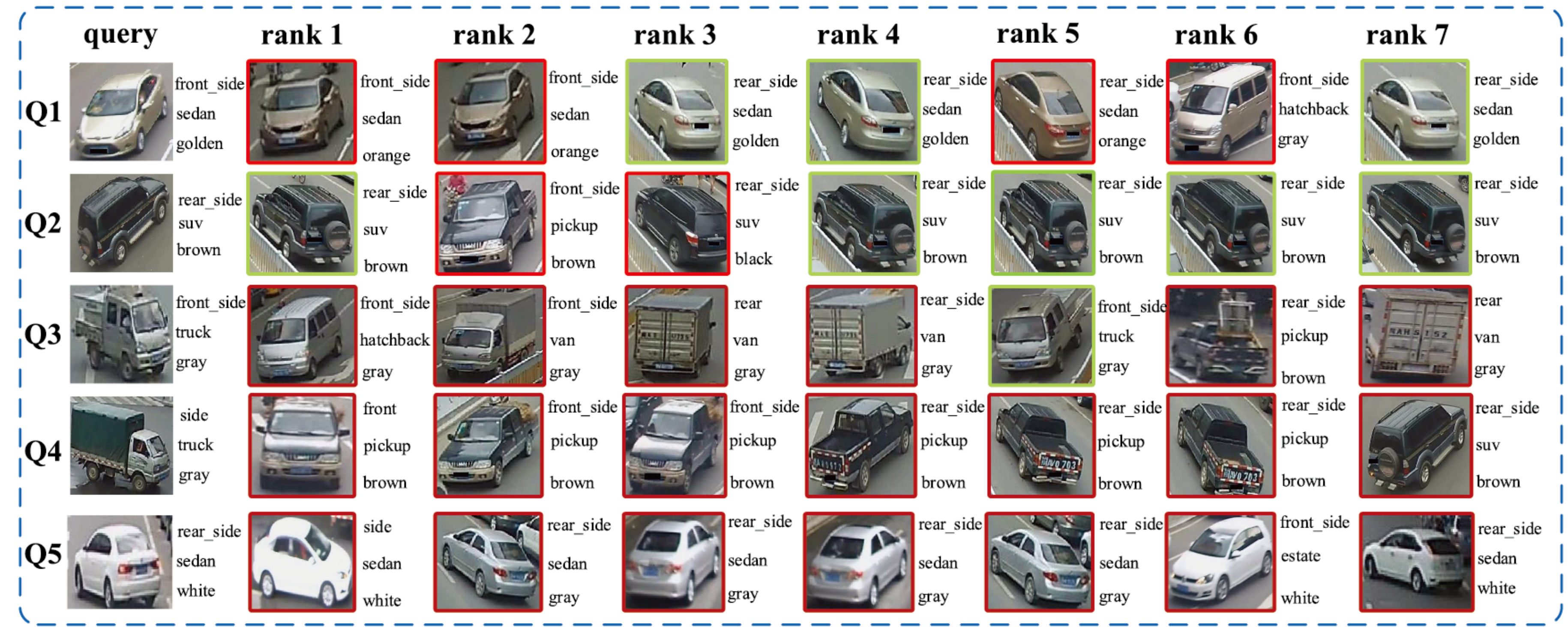}
 \caption{Benefits of camera views, types and colors on vehicle Re-ID. The {\color{blue}blue dash} box demonstrates several ranking results of conventional vehicle Re-ID based on ResNet-50~\cite{he2016deep}, where the {\color{red}red} and {\color{green}green} solid boxes of the first 7 ranks indicate the wrong and right matching respectively. The results show that extra semantics or attributes play critical role in handling the challenges of vehicle Re-ID. }\label{fig::motivation}
\end{figure*}

The second challenge is the high inter-class similarity especially for different identities with a similar appearance. Incorporating the attributes information suffices to generate better discriminative representation for person Re-ID~\cite{su2015multi,su2016deep,khamis2014joint,layne2012person}.
Therefore, it is essential to learn the deep features with the supervision of attributes in vehicle Re-ID, enforcing the same identity with the consistent attributes. Li et al.~\cite{li2017deep} introduce the attribute recognition into the vehicle Re-ID framework, and use extra semantic information to assist vehicle identification especially for different identities with a similar appearance.
Qian et al.~\cite{qian2020stripe} propose a two-branch stripe-based and attribute-aware deep convolutional neural network (SAN), in which attribute information and part-level features are combined to enhance the discriminative capability for vehicle Re-ID.
However, none of the methods handles both of two challenges (intra-class difference and inter-class similarity) simultaneously.

Motivated by the human visual system that recognizes the vehicle by progressively identifying the color, type with various viewpoint, we propose a unified deep convolutional framework to learn Deep Feature representations jointly guided by the meaningful attributes, including Camera Views, vehicle Types and Colors (DF-CVTC) for vehicle Re-ID.
Attribute information has been successfully investigated as the mid-level semantics to boost person Re-ID. It can also help vehicle Re-ID in challenging scenarios.
First of all, the camera view is one of the key attributes and challenges in Re-ID. As shown in Fig.~\ref{fig::motivation}, the query vehicle image may have completely different views from their counterparts under other cameras, such as query Q1 and Q2 and their right ranks marked as green solid boxes.
Second, vehicle types and colors, as the representative attributes for vehicles, also play an important role in vehicle Re-ID especially for the different vehicles with a similar appearance.
As shown in Fig.~\ref{fig::motivation}, the wrong hits of query Q3 and Q4, which present with similar appearance, could be effectively evaded by the vehicle type.
Furthermore, the vehicles with different colors may present with similar shapes (such as the wrong hits rank 1 and rank 2 of query Q1), similar overall appearance (such as the wrong hits of query Q4), or even the similar color (such as the query Q5 with white color while the wrong hits of rank 2-4 with gray color) due to illumination changes.
Integrating the color attribute may relieve this inter-class similarity. These challenges motivate us to utilize the above attributes to help network distinguish different vehicles with very similar appearance and also identify the same vehicles with different viewpoints.


It is worth noting that we jointly learn deep features, camera views, vehicle types and colors in an end-to-end framework. At last, we annotate the view labels in the benchmark datasets for network training, which can be directly used for other datasets with only ID labels.
Comprehensive evaluations on two benchmark datasets, i.e., VeRi-776 and VehicleID, demonstrate the promising performance of the proposed method which yields to a new state-of-the-art for vehicle Re-ID.

In summary, this paper makes the following contributions to vehicle Re-ID and related applications:
\begin{itemize}
  \item It proposes a unified attributes guided deep learning framework that jointly learns Deep Feature representations, Camera Views, vehicle Types and Colors (DF-CVTC) for vehicle Re-ID.
%
These components are collaborative to each other, and thus boost the performance of vehicle Re-ID significantly.


\item We annotate the view labels for the benchmark dataset VeRi-776 for view predictor training, which could be easily employed for the situation with only ID information available in vehicle Re-ID. We will release the annotation information of view labels to the public for free academic usage~\footnote{http://www.escience.cn/people/AihuaZheng/Code-Dataset.html}.

\end{itemize}

\section{Related Work}

\subsection{Vehicle Re-ID}
With great progress in person Re-ID~\cite{zhong2018camera,zhu2018fast,wang2019pedestrian}, vehicle Re-ID has gradually gained a lot of attention recently since vehicles are the most important object in urban surveillance.
Liu et al.~\cite{liu2016large,liu2016deep,liu2018provid} released a benchmark dataset VeRi-776 and considered the vehicle Re-ID task as a progressive recognition process by using visual features, license plates and spatial-temporal information.
Liu et al.~\cite{liu2016deep} released another big surveillance-nature dataset (VehicleID) and designed coupled clusters loss to measure the distance of two arbitrary similar vehicles.
Tang et al.~\cite{tang2019cityflow} introduced a dataset CityFlow, which is currently the largest-scale dataset in terms of spatial coverage and the number of cameras/videos in an urban environment.
Lou et al.~\cite{lou2019veri} collected a new dataset called VERI-Wild for vehicle Re-ID community in the wild, and designed a novel feature distance adversary scheme to online generate hard negative samples in feature space to facilitate Re-ID model training.
He et al.~\cite{he2019part} developed a new end-to-end framework to integrate part constrains with the global
Re-ID modules by introducing an detection branch.
Zhang et al.~\cite{zhang2017improving} designed an improved triplet-wise training by classification-oriented loss. Li et al.~\cite{li2017deep} integrated the identification, attribute recognition, verification and triplet tasks into a unified CNN framework.
Liu et al.~\cite{liu2018learning} proposed a coarse-to-fine ranking method consisting of a vehicle model classification loss, a coarse-grained ranking loss, a fine-grained ranking loss and a pairwise loss.

In addition to appearance information, Shen et al.~\cite{shen2017learning} combined the visual spatio-temporal path information for regularization.
%
Hsu et al.~\cite{hsu2019multi} exploited the temporal attention model to extract the most discriminant feature of each trajectory.
Chen et al.~\cite{chen2020orientation} proposed a dedicated Semantics-guided Part Attention Network (SPAN) to robustly predict part attention masks for different views of vehicles given only image-level semantic labels during training.
However, the large intra-class variation and inter-class similarity in different viewpoints have not been well studied in existing works. In this paper, we propose to embed the viewpoint as well as the attributes information into the appearance information for better discriminative feature learning.

\subsection{View-aware Re-ID}
Viewpoint changes introduce a large variation of the intra-class variation in person Re-ID.
Zhao et al.~\cite{zhao2017spindle} proposed a novel method based on human body region guided for person Re-ID which can boost the performance well.
Wu et al.~\cite{wu2015viewpoint} proposed an approach called pose prior to make identification more robust to the viewpoint.
Zheng~\cite{zheng2019pose} introduced the PoseBox structure which is generated through pose estimation followed by affine transformations.
Qian et al.~\cite{qian2018pose} use GAN to generate eight pre-defined pose for each image which augment the data and address the viewpoint variation to some extent.
Liu et al.~\cite{liu2018pose} transferred various person pose instances from one dataset to another to improve the generalization ability of the model.
Zhou et al.~\cite{zhou2018cross} designed a conditional generative network to obtain cross-view images from input view pairs to address the vehicle Re-ID task.
Later on, Zhou et al.~\cite{zhou2018aware} proposed a Viewpoint-aware Attentive Multi-view Inference (VAMI) model to infer multi-view features from single-view image inputs.

This issue is even crucial in vehicle Re-ID, since the viewpoint of the images are almost the same due to the rigid motion of the vehicles.
Prokaj et al.~\cite{prokaj20093} presented a method based on pose estimation to deal with multiple viewpoints.
However, different vehicle identities might present a similar appearance while th auxiliary attributes information could help to distinguish them. Therefore, we propose to integrate several attributes information into a joint deep feature learning framework in this paper.

%

\subsection{Attribute Embedded Re-ID}
Attributes have been extensively investigated as the mid-level semantic information to boost the person Re-ID.
Su et al.~\cite{su2015multi} introduced a low rank attribute embedding into the multi-task learning framework for person Re-ID.
Khamis et al.~\cite{khamis2014joint} jointly optimized the attributes classification loss and triplet loss for person Re-ID.
Lin et al.~\cite{lin2017improving} integrated the identification loss and the attributes prediction into a simple ResNet framework and annotated the pedestrian attributes in two benchmark person Re-ID datasets Market-1501 and DukeMTMC-reID.
Su et al.~\cite{su2018multi} proposed a weakly supervised multi-type attribute learning framework based on the triplet loss by pre-training the attributes predictor on independent data.
Despite the previous works focusing on image-based query, Li et al~\cite{li2017person} and Yin et al.~\cite{yin2018adversarial} investigated attribute-based query for person retrieval and Re-ID task.

In vehicle Re-ID, Li et al.~\cite{li2017deep} introduced the attribute recognition into the vehicle Re-ID framework together with the verification loss and triplet loss.
Qian et al.~\cite{qian2020stripe} proposed a novel two-branch stripe-based and attribute-aware deep convolutional neural network (SAN) to learn the efficient feature embedding for vehicle Re-ID task.
Different from these methods, we take the view-aware identification and attributes recognition into a unified vehicle Re-ID framework.

\begin{figure*}[t]
\centering
\includegraphics[width =0.9\textwidth]{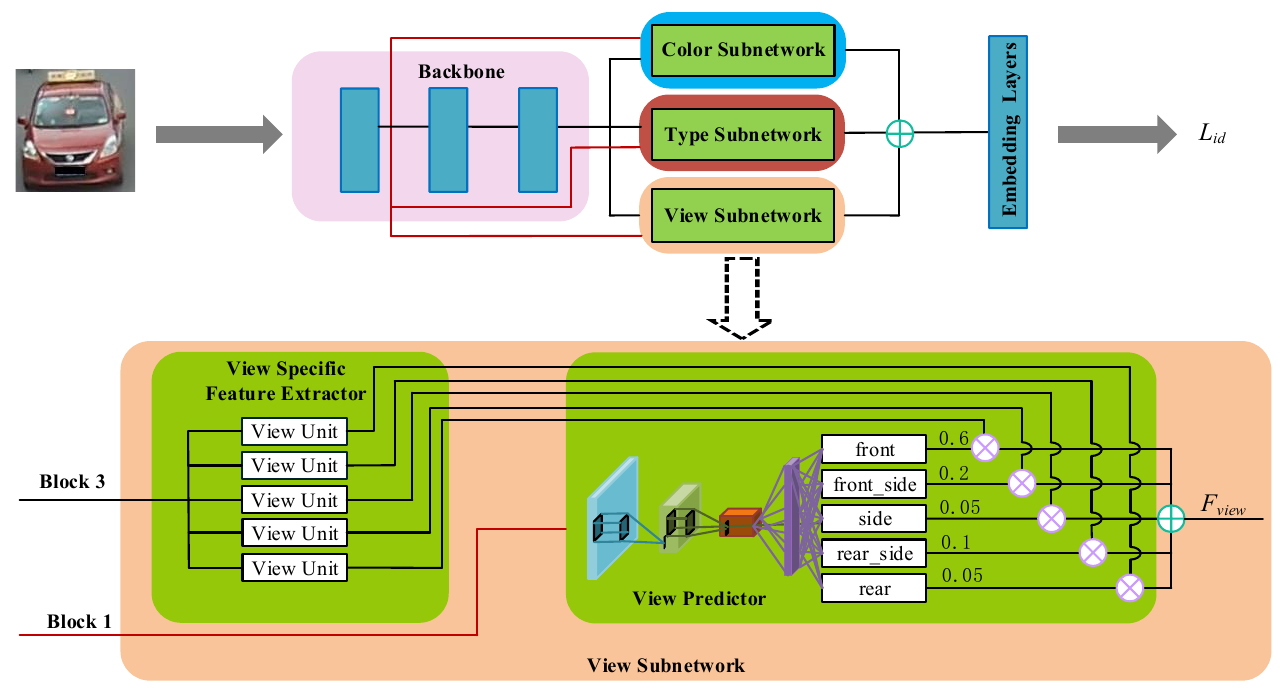}
\caption{Overview of our DF-CVTC, which consists of one backbone (the first three blocks of ResNet-50),  three subnetworks and one embedding network.}\label{fig::framework}
\end{figure*}

\section {DF-CVTC Network Architecture}
 In this paper, we propose a novel Deep Feature learning method, which embeds attributes information, including Camera views, Vehicle Types and Colors (DF-CVTC), for vehicle Re-ID. We shall elaborate the proposed method in this section.

\subsection {Architecture Overview}
The overall architecture is demonstrated in Fig.~\ref{fig::framework}.
To explore and make full use of the auxiliary information of the vehicle, we firstly employ a backbone to learn the shared features, followed by three subnetworks to extract the corresponding attribute weighted feature specifically.
Finally, we use an embedding layer to generate the attribute embedded features for vehicle Re-ID. We shall elaborate our model in the following of this section.

\subsection {Backbone}
\label{sec:backbone}
Due to the compelling performance with deeper layers by residual learning, ResNet-50 has been widely used in many research~\cite{liu2018segmentation,ding2019neural}. Therefore, we adopt the first three residual blocks of ResNet-50 as the baseline network for our backbone as shown in Fig.~\ref{fig::framework}.
One could also configure other networks such as Inception-v4 \cite{szegedy2017inception}, VGG16 \cite{Simonyan15} and MobileNet \cite{howard2017mobilenets} architectures without limitation.
The task of each attribute recognition shares the low-level properties which could be efficiently learned by the shared backbone.
%

\subsection {Subnetworks}
As shown in Fig.~\ref{fig::framework}, each subnetwork consists of a predictor part and a feature extraction part.
To share the low-level information and reduce the complexity of our network, we use the feature of Block-1 as the input of the following predictor part.

The predictor is composed of  three  convolutional (Conv) layers and one fully-connected (FC) layer which outputs a probability distribution over the corresponding (view, type or color) values.
The kernel sizes in the three Conv layers are $5\times5$, $3\times3$, $5\times5$, respectively. The strides for these kernels are 3, 2 and 1, respectively.
We use ReLU activation in all three layers and add a $batch$ $normalization$ layer after each Conv layer.
The resulting feature vector is fed into the following FC layer to predict the attribute scores via the $K$-way softmax.

The feature extractor is composed of $K$ units, each of which is a Conv-net responsible for extracting high-level features corresponding to one of the $K$ view or attribute classes. We use the  Block-4 of ResNet-50 as feature extractor.

The features from each specific feature extractor $E_{\Phi}$ can be formulated as,
\begin{equation}
\begin{split}
f_{{\Phi}_k}=E_{\Phi}(x;\alpha_{\Phi})
\end{split}
\end{equation}
where $\Phi \in \{view ,type, color\}$, $k = 1, 2, ..., K_{\Phi}$. $K_{\Phi}$ is the number of corresponding units, which also indicates the possible classes of each view or attribute. $x$ is an image, $\alpha_{\Phi}$ denotes the parameters of $E_{\Phi}$.

The probability distribution $w_{\Phi}$ over corresponding view or attribute values from the predictor network $P_{\Phi}$ is,
\begin{equation}
\begin{split}
 w_{\Phi}=P_{\Phi}(x;\beta_{\Phi})
\end{split}
\end{equation}
where $\beta_{\Phi}$ denotes the parameters of $P_{\Phi}$, which is learnt using the cross-entropy loss $\mathcal{L}_{\Phi}$,
\begin{equation}
\mathcal{L}_{\Phi} = -\sum_{k=1}^{K_{\Phi}}log( w_{\Phi}(k))q_{\Phi}(k)
\end{equation}
where $q_{\Phi}$ is a one-hot vector of the ground truth of corresponding view or attributes values.

After progressively learning of the three subnetworks, we achieve the specific feature maps via:
\begin{equation}
\begin{split}
F_{\Phi}=(f_{{\Phi}_1} \odot w_{{\Phi}_1}) \oplus ... \oplus(f_{{\Phi}_K} \odot w_{{\Phi}_K})
\end{split}
\end{equation}
where $\oplus$ denotes the element-wise sum, and $\odot$ denotes the element-wise multiply.
$F_{\Phi}$ is the augmented features by element-wise sum operation in each channel dimension, which can be beneficial for classification without extra computation.
The joint deep features with camera view, type and color are achieved as the fusion of feature maps of three subnetworks,
\begin{equation}
\begin{split}
F= F_{view} \oplus F_{type} \oplus F_{color}\label{eq::fusionF}
\end{split}
\end{equation}
$F$ is the fused deep features containing the complementary view and attributes information.
The feature dimensions of $F_{view}$, $F_{type}$, and $F_{color}$ are all $2048$.
Next, we describe the details of each subnetwork as follows.

\subsubsection {View subnetwork}
Viewpoint changes bring a crucial challenge for the Re-ID task.
We use the view subnetwork to incorporate the view information into the Re-ID model.
The view predictor predicts $K_{view}$-way softmax scores which are used to weight the output of each corresponding view unit.
Followed by \cite{zhou2018aware}, we horizontally flip each vehicle image, which means we do not distinguish left and right viewpoints. Therefore the vehicle images fall into five viewpoints in this paper: $front, front\_side, side, rear\_side$, and $rear$.

For instance, for the training sample in the rear orientation to the camera, the corresponding view unit will be assigned a strong weight and updated strongly during the back propagation. 

\subsubsection {Type subnetwork}
 Type is useful to distinguish the vehicles with similar appearance, which can relieve the inter-class similarity.
In the same manner as the view subnetwork, we use the type subnetwork to learn the attribute specific deep features.
The $K_{type}$ scores predicted by type predictor are used to weight the output of each corresponding type unit in the type specific feature extractor.
Following the protocol of type attribute information in VeRi-776 dataset~\cite{liu2016large}, in this paper, we set $K_{type}$ = 9, indicating 9 types of the vehicles: $sedan$, $suv$, $van$, $hatchback$, $mpv$, $pickup$, $bus$, $truck$ and $estate$.
Similar to previous view specific feature extractor, each type unit will learn a feature map specialized for one of the $K_{type}$ types. 

\subsubsection {Color subnetwork}
Color is another discriminative attribute in vehicles.
Therefore, we analogously use the color subnetwork to learn the color-specific features.
The color predictor predicts the color scores of the vehicle then weight to each color unit.
In the same manner as the color attribute is given in VeRi-776 dataset~\cite{liu2016large}, in our implementation, we set $K_{color}$ =10 denoting 10 colors of the vehicles: $yellow$, $orange$, $green$, $gray$, $red$, $blue$, $white$, $golden$, $brown$ and $black$.
The color-specific feature extractor is designed in the same manner as in view and type subnetworks.

\subsection {Embedding Layers}
The embedding layers consist of two FC layers.
It embeds the fused feature $F$ in Eq.~\eqref{eq::fusionF} into the higher level joint deep feature $F_{joint}$, which is used for the final Re-ID task.

In order to train the Re-ID model, we add a softmax layer into the embedding network for ID classification.
We use the cross-entropy loss of $\mathcal{L}_{id}$ for model training,
\begin{equation}
\mathcal{L}_{id} = -\sum_{n=1}^Nlog(p_{id}(n))q_{id}(n)
\end{equation}
where $N$ is the number of the vehicle IDs in the training set. $q_{id}$ is the one-hot ground-truth of the ID label of the vehicle. $p_{id}(n)$ is the predicted probability indicating the ID of the input vehicle image,
\begin{equation}
{p}_{id} = softmax(F_{joint}).
\end{equation}

\begin{figure}[t]
\includegraphics[width =0.50\textwidth]{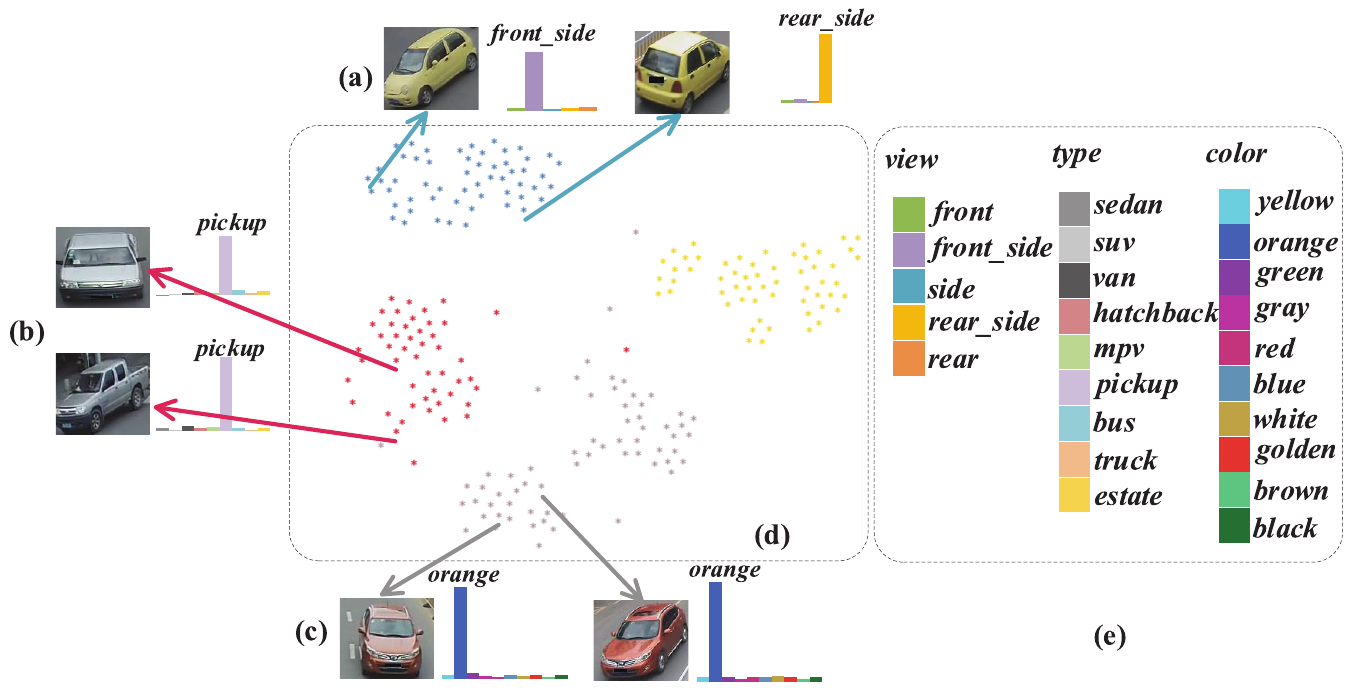}
 \caption{Demonstration of the DF-CVTC features. (a) (b) and (c) denote  three vehicle pairs sampled from VeRi-776 dataset under two distinct camera views and their corresponding learned probabilities of varying camera views, types and colors, where visible appearances are distinct due to the different camera views or illumination changes. (d) illustrates the 2D feature projections of the vehicle images learnt by the proposed DF-CVTC. (e) represents the corresponding annotation categories of camera views, types, and colors.}\label{fig::featurespace}
\end{figure}

Fig.~\ref{fig::featurespace} demonstrates the effectiveness of the jointly learnt deep features of the  proposed DF-CVTC.
We can observe that, the vehicle images of the same identity fall into the same cluster regardless of the different visible appearance caused by different camera views (as shown in Fig.~\ref{fig::featurespace}(a) and (b)) or illumination changes (as shown in Fig.~\ref{fig::featurespace} (c)).

\subsection{Difference from Previous Work}
Our method is significantly different from~\cite{zhou2018cross,zhou2018aware,saquib2018pose} from the following aspects. First,~\cite{zhou2018cross,zhou2018aware} infer the multi-view images or features using adversarial learning. However, they render vehicle Re-ID as a verification task while our method employs a classification CNN to learn the deep features. Furthermore, our learnt features embeds attributes information (type and color) in addition to the view information. Second,~\cite{saquib2018pose} incorporate both fine and coarse pose/view information to learn a feature representations and propose a novel re-ranking method for person Re-ID. While our DF-CVTC further integrates the attributes information and jointly learns the deep features embedded by camera views, vehicle types and colors into an end-to-end framework.

\section{Training Details}

\subsection{Progressive Learning}
We progressively learn the three subnetworks and fine-tune the DF-CVTC model, which achieves comparative performance as the multi-task learning (minimizing the combination of the four losses).
Furthermore, it can significantly reduce the computational complexity.

\subsubsection{View subnetwork training}
We fine-tune the backbone network pre-trained on ImageNet classification~\cite{krizhevsky2012imagenet} and the rest of Re-ID model are initialized from scratch. %
First, we minimize  $\mathcal{L}_{view}$ to obtain ${\bf \alpha}_{view}$, then we minimize  $\mathcal{L}_{id}$ to obtain $\{ {\bf \beta}_{view}, { \theta_2}\}$ while fixing all the other parameters in Re-ID model.

\subsubsection{Type subnetwork training}
We first minimize  $\mathcal{L}_{type}$ to obtain ${\bf \alpha}_{type}$, and then minimize  $\mathcal{L}_{id}$ using $F_{view} \oplus F_{type}$ to obtain $\{ {\bf \beta}_{type}, {\theta_2}\}$ while fixing all the other parameters in Re-ID model.

\subsubsection{Color subnetwork training}
In the same manner, we first minimize  $\mathcal{L}_{color}$ to obtain ${\bf \alpha}_{color}$, then minimize  $\mathcal{L}_{id}$ using $F_{view} \oplus F_{type} \oplus F_{color}$ to obtain $\{ {\bf \beta}_{color}, { \theta_2}\}$ while fixing all the other parameters in Re-ID model.

\subsubsection{Joint learning}
After training the three subnetworks, we fine-tune $\{ {\bf \alpha}_{\Phi}, {\bf \beta}_{\Phi}, { \theta_1}, { \theta_2}\}$, $\Phi \in \{view ,type, color\}$ of the whole Re-ID model by minimizing $\mathcal{L}_{id}$ until convergence.

\subsection{Implementation Details}
In practice, we use a stochastic approximation of the objective since the training set is quite huge.
The training set is stochastically divided into mini-batches with 16 samples. The network performs forward propagation on the current mini-batch, followed by the backpropagation to compute the gradients with simple cross-entropy loss for network parameters updating.
We perform Adam optimizer at recommended parameters with an initial learning rate of 0.0001 and a decay of 0.96 every epoch.
With more passes over the training data, the model improves until it converges. To reduce overfitting, we artificially augment the data by performing random 2D translation as the same protocol in ~\cite{li2014deepreid}.
In our implementation, all the input images are resized to $W \times H = 256 \times 256$.

\begin{figure*}[!t]
\centering
\includegraphics[width =0.9\textwidth]{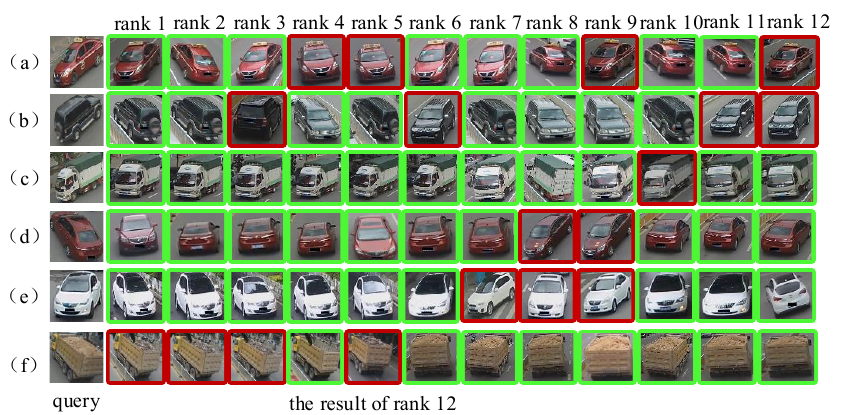}
\caption{Examples of ranking results on VeRi-776 dataset. The {\color{green} green} and {\color{red}red} boxes indicate the right matchings and the wrong matchings respectively.}\label{fig::rankings}
\end{figure*}

\section{Experiments}
We carry out a comprehensive evaluation of the proposed DF-CVTC comparing to the state-of-the-art methods on two public vehicle Re-ID datasets, VeRi-776~\cite{liu2016large} and VehicleID~\cite{liu2016deep}.
We use the Cumulative Matching Characteristics (CMC) curves and mAP to evaluate our results~\cite{liu2016deep}.
The type and color labels are available in VeRi-776, therefore, we annotate the view labels for network training.
In VehicleID, we directly employ the view, type and color subnetworks pre-trained on VeRi-776 and only ID labels are used.  
\begin{figure*}[!t]
\centering
\includegraphics[width =0.9\textwidth]{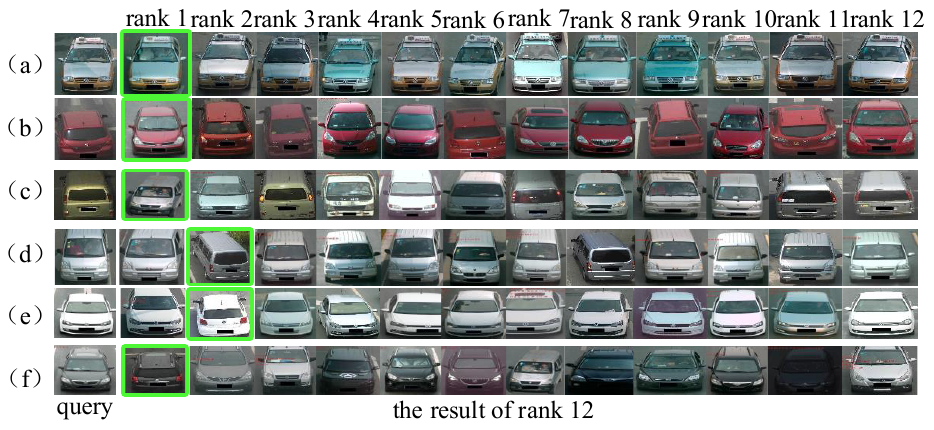}
\caption{Examples of ranking results on VehicleID dataset. The {\color{green} green} boxes indicate the right matchings. Note that there is only one ground truth vehicle image in gallery set in the VehicleID dataset.}\label{fig::rankings2}
\end{figure*}

\begin{table}
\caption{ Comparisons with state-of-the-art Re-ID methods on VeRi-776 (in \%). The top three results are highlighted in {\color{red} red}, {\color{green} green} and {\color{blue} blue}, respectively.}\label{tb::veri776}
\begin{center}
\setlength{\tabcolsep}{3.2 mm}{

\begin{tabular}{l|ccc}
\hline
Method     & mAP & rank 1 & rank 5 \\
\hline
(1) LOMO~\cite{liao2015person}       & 9.64 &25.33 &46.48   \\
(2) BOW-CN~\cite{zheng2015scalable}        & 12.20 &33.91 &53.69    \\
(3) GoogLeNet~\cite{yang2015large}        & 17.89 &52.32 &72.17    \\
(4) FACT~\cite{liu2016large}  & 18.49 & 50.95 &73.48    \\
(5) FACT+Plate-SNN+STR~\cite{liu2016deep} &27.70 &61.44 &78.78  \\
(6) Siamese-Visual~\cite{shen2017learning} &29.48 &41.12 &60.31  \\
(7) Siamese-CNN+Path-LSTM~\cite{shen2017learning} &58.27 &83.49 &90.04  \\
(8) NuFACT~\cite{liu2018provid} &48.47 &76.76 &91.42  \\
(9) VAMI~\cite{zhou2018aware} &50.13 &77.03 &90.82  \\
(11) EALN \cite{lou2019embedding} &57.44 &84.39 &94.05 \\
(12) AAVER~\cite{khorramshahi2019dual} &{\bf{\color{blue}58.52}} &{\bf{\color{green}88.68}} &{\bf{\color{blue}94.10}} \\
(13) VFL~\cite{alfasly2019variational} &{\bf{\color{green}59.18}} &{\bf{\color{blue}88.08}} &{\bf{\color{green}94.63}} \\

\hline
~~DF-CVTC &{\bf{\color{red} 61.06}} &{\bf {\color{red}91.36}} &{\bf {\color{red}95.77}} \\
\hline
\end{tabular}}
\end{center}
\end{table}

\begin{table*}[t]
\caption{Comparisons with state-of-the-art Re-ID methods on VehicleID (in \%). The top three results are highlighted in {\color{red} red}, {\color{green} green} and {\color{blue} blue}, respectively.}\label{tb::VehicleID}
\begin{center}
\setlength{\tabcolsep}{4.2mm}{
\begin{tabular}{l|ccc|ccc|ccc}
\hline
 \multirow{2}*{Method} &\multicolumn{3}{c|}{\textbf{\emph{Test Size = 800}}} &\multicolumn{3}{c|}{\textbf{\emph{Test Size = 1600}}} &\multicolumn{3}{c}{\textbf{\emph{Test Size = 2400}}} \\
 &mAP &rank 1 & rank 5  &mAP & rank 1 & rank 5 &mAP & rank 1 & rank 5 \\
\hline
(1) LOMO~\cite{liao2015person} &- & 19.76 &32.01 &- &18.85 &29.18 &- &15.32 &25.29  \\
(2) BOW-CN~\cite{zheng2015scalable} &- &13.14 &22.69 &- &12.94 &21.09 &- &10.20 &17.89  \\
(3) GoogLeNet~\cite{yang2015large} &46.20 &47.88 &67.18 &44.00 &43.40 &63.86 &38.10 &38.27 &59.39   \\
(4) FACT~\cite{liu2016large} &- &49.53 &68.07 &- & 44.59 &64.57 &-&39.92 &60.32 \\
(8) NuFACT~\cite{liu2018provid} &-&48.90 &69.51 &-&43.64 &65.34 &-&38.63 &60.72 \\
(9) VAMI~\cite{zhou2018aware} &- &63.12 &83.25 &- &52.87 &75.12 &- &47.34 &70.29 \\
(10) C2F-Rank~\cite{liu2018learning}  &{\bf{\color{blue}63.50}} &61.10 &81.70 &{\bf{\color{blue}60.00}} &56.20 &76.20 &{\bf{\color{blue}53.00}} &51.40 &72.20 \\

(11) EALN~\cite{lou2019embedding}
&{\bf{\color{green}77.50}} &{\bf{\color{green}75.11}} &{\bf{\color{green}88.09}} &{\bf{\color{green}74.20}} &{\bf{\color{green}71.78}} &{\bf{\color{green}83.94}} &{\bf{\color{green}71.00}} &{\bf{\color{green}69.30}} &{\bf{\color{green}81.42}} \\

(13) VFL~\cite{alfasly2019variational}
&- &{\bf{\color{blue}73.37}} &{\bf{\color{blue}85.52}} &- &{\bf{\color{blue}69.52}} &{\bf{\color{blue}81.00}} &- &{\bf{\color{blue}67.41}} &{\bf{\color{blue}78.48}} \\

\hline
~~DF-CVTC &{\bf {\color{red}78.03}} &{\bf {\color{red}75.23}} &{\bf {\color{red}88.11}} &{\bf {\color{red}74.87}} &{\bf {\color{red}72.15}} &{\bf {\color{red}84.37}} &{\bf {\color{red}73.15}} &{\bf {\color{red}70.46}} &{\bf {\color{red}82.13}}  \\
\hline
\end{tabular}}
\end{center}
\end{table*}

\begin{table*}[t]
\caption{Influence of different component on VehicleID and VeRi-776 dataset (in \%). The top three results are highlighted in {\color{red} red}, {\color{green} green} and {\color{blue} blue}, respectively. }\label{tb::component}
\begin{center}
\setlength{\tabcolsep}{1.6mm}{
\begin{tabular}{l|ccc|ccc|ccc|ccc}
\hline
&\multicolumn{9}{c|}{\bf VehicleID} &\multicolumn{3}{c}{\bf VeRi-776}\\
\hline
 \multirow{2}*{Method} &\multicolumn{3}{c|}{\textbf{\emph{Test Size = 800}}} &\multicolumn{3}{c|}{\textbf{\emph{Test Size = 1600}}} &\multicolumn{3}{c|}{\textbf{\emph{Test Size = 2400}}} \\
 &mAP &rank 1 & rank 5  &mAP & rank 1 & rank 5 &mAP & rank 1 & rank 5 &mAP & rank 1 & rank 5\\
\hline
ResNet-50 (baseline)  &70.50 &67.75 &79.13 &68.48&65.79 &76.64 &66.19&63.45 &74.70 &51.58 &86.71 &92.43    \\
~~+view  &{\color{blue}75.44} &{\color{blue}72.63} &{\color{blue}84.82} &{\color{blue}72.41} &{\color{blue}69.62} &{\color{blue}81.36} &{\color{blue}70.71} &{\color{blue}68.02} &{\color{blue}79.19} &{\color{blue}54.52} &{\color{blue}89.69} &{\color{blue}94.40} \\
~~+view+type  &{\color{green}76.06} &{\color{green}73.14} &{\color{green}86.25} &{\color{green}73.39} &{\color{green}70.77} &{\color{green}81.75} &{\color{green}71.75} &{\color{green}69.10} &{\color{green}80.40} &{\color{green}60.47} &{\color{red}91.66} &{\color{green}95.59}\\
~~+view+type+color (DF-CVTC) &{\color{red}78.03} &{\color{red}75.23} &{\color{red}88.11} &{\color{red}74.87} &{\color{red}72.15} &{\color{red}84.37} &{\color{red}73.15} &{\color{red}70.46} &{\color{red}82.13} &{\color{red}61.06} &{\color{green}91.36} &{\color{red}95.77}\\
\hline
\end{tabular}}
\end{center}
\end{table*}

\subsection{Experiments on VeRi-776 Dataset}
\subsubsection{Setting}
The VeRi-776 dataset contains 776 identities collected with 20 cameras in a real-world traffic surveillance environments.
The whole dataset is split into 576 identities with 37,778 images for training and 200 identities with 11,579 images for testing.
An additional set of 1,678 images selected from the test identities are used as query images.
In order to evaluate the view subnetwork, we annotate all the vehicle images in VeRi-776 dataset into five viewpoints as $front$, $front\_side$, $side$, $rear\_side$ and $rear$.
We follow the evaluation protocol in~\cite{liu2016deep}. We use mean average precision (mAP) metric for evaluation. We first calculate the average precision for each query.
Then, the mAP can be obtained by calculating the mean of each average precision. The cumulative match curve (CMC) metric is also used for evaluation.
First, we sort the Euclidean distance between each query and gallery images in ascending order.
Then, the CMC curve can be obtained by the average of sorted value. Noted that, only the vehicles in non-overlap cameras are counted during evaluation.

\subsubsection{Qualitative examples}
Fig.~\ref{fig::rankings} demonstrates the qualitative examples of six ranking results of our DF-CVTC on VeRi-776 dataset.
From which we can observe that our method successfully hits the vehicles with large view variations to the query such as rank 2 and ranks 10-11 in Fig.~\ref{fig::rankings} (a), rank 4 and ranks 8-9 in Fig.~\ref{fig::rankings} (b), rank 8 in Fig.~\ref{fig::rankings} (c), rank 1 and rank 5 in Fig.~\ref{fig::rankings} (d).
The wrong hits generally result from the high inter-class similarity with homologous visual appearance, such as ranks 4-5, rank 9 and rank 12 in Fig.~\ref{fig::rankings} (a), rank 10 in Fig.~\ref{fig::rankings} (c), ranks 7-9 in Fig.~\ref{fig::rankings} (e), ranks 1-3 and rank 5 in Fig.~\ref{fig::rankings} (f).
Fig.~\ref{fig::progress} demonstrates the qualitative examples of ranking result of our DF-CVTC on VeRi-776 dataset.
Fig.~\ref{fig::progress} (b) shows the view/attributes probability which is predicted by each subnetwork. Fig.~\ref{fig::progress} (c) show the ranking result.
From the Fig.~\ref{fig::progress}, we can find that the ranking result is improving by introduction each subnetwork progressively.

From the above observation, we can conclude that our method can reduce the influence of viewpoint changes to some extends.

\subsubsection{Quantitative results}
Table~\ref{tb::veri776} reports the performance of our approach comparing with the published state-of-the-arts on VeRi-776 dataset.
From which we can see, our DF-CVTC significantly surpasses the state-of-the-art.
Compared with the second best method VFL~\cite{alfasly2019variational}, our method achieves 1.88\% and 3.28\% improvements in terms of mAP and rank 1 respectively.
Note that we haven't utilized any license plates or spatial temporal information as in Siamese-CNN+Path-LSTM~\cite{shen2017learning} and FACT+Plate-SNN+STR~\cite{liu2016deep}.
Even though, our method still achieves the superior mAP and ranking accuracies by a large margin. Such a huge improvement verifies the effectiveness of our proposed method for the vehicle Re-ID.


\subsection{Experiments on VehicleID Dataset}
\subsubsection{Setting}
 The VehicleID dataset~\cite{liu2016deep} consists of the training set with 110,178 images of 13,134 vehicles and the test set with 111,585 images of 13,133 vehicles.
Followed by the protocol in~\cite{liu2016deep}, we test VehicleID dataset in three distinct settings with different number of testing samples: 800, 1600  and 2400.
Specifically, since some of the type and color information is missing and no view labels in this dataset, we adopt the view, type and color subnetworks pre-trained on VeRi-776 dataset and fine-tuned during the Re-ID training.
Which in turn means one can easily apply our model on the dataset with only ID information. The mean average precision (mAP), cumulative match curve (CMC) are used as the evaluation metric in the same manner as in VeRi-776.
The only difference is we randomly select a image from test dataset as gallery, while consider the remaining images in test dataset as query. The experimental results are based on the average of 10 random trials.

\subsubsection{Qualitative examples}
Fig.~\ref{fig::rankings2} demonstrates six ranking results of our DF-CVTC on VehicleID.
From which we can observe that, our method can successfully hit the right matching with large inter-class difference caused by the illumination/color changes, such as Fig.~\ref{fig::rankings2} (a), (c) and (f), as well as the viewpoint changes, such as Fig.~\ref{fig::rankings2} (b), (c), (d) and (f).
The wrong hits of rank 1 on Fig.~\ref{fig::rankings2} (d) and (e) result from the inter-class similarity between vehicles, despite of which, our method still hit the right matchings in the early ranks.
Note that there is only one ground truth vehicle image in gallery set in the VehicleID dataset.

We can easily draw the conclusion that our method tends to hit the vehicles with the same type and color despite the change of viewpoints.

\subsubsection{Quantitative results}
Table~\ref{tb::VehicleID} reports the performance of our method against the state-of-the-arts on VehicleID dataset.
Clearly, our method significantly beats the existing state-of-the-arts in mAP, rank 1 and rank 5.
Based on above results, we can conclude that our method achieves significant improvement for vehicle Re-ID.


\subsection{Ablation Study}

\begin{figure*}[]
\centering
\includegraphics[width =0.8\textwidth]{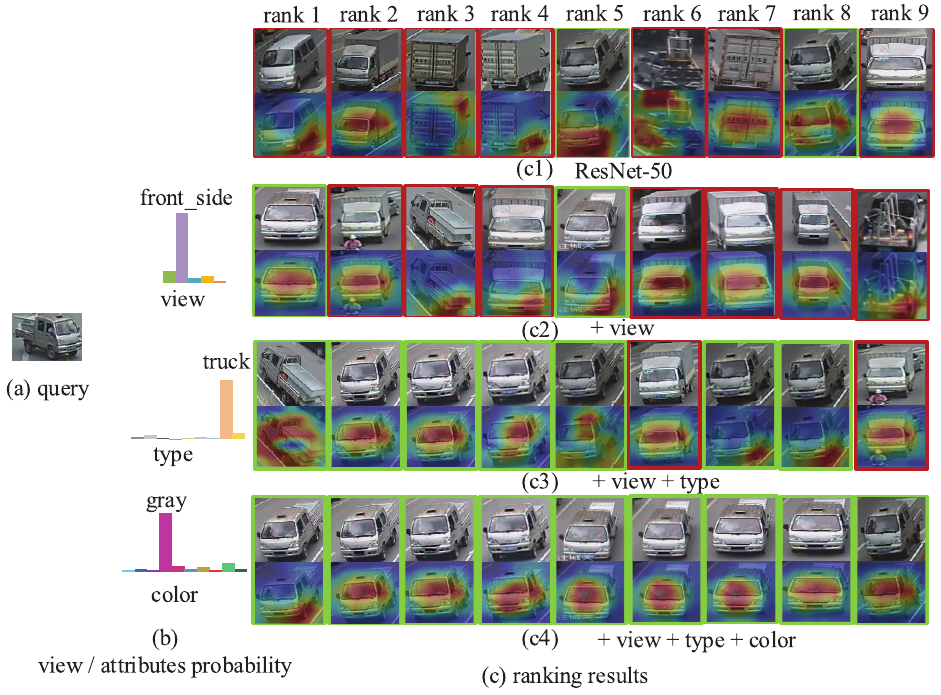}
\caption{An example of ranking results together with corresponding CAM visualizations of DF-CVTC on ResNet-50 backbone by progressively introducing the view, type and color subnetworks on VeRi-776 dataset. The {\color{green} green} and {\color{red}red} boxes indicate the right matchings and the wrong matchings respectively. The histograms denote the probability distributions learnt from the view, type and color subnetworks respectively. }\label{fig::progress}
\end{figure*}

\begin{figure}[]
\centering
\includegraphics[width =0.51\textwidth]{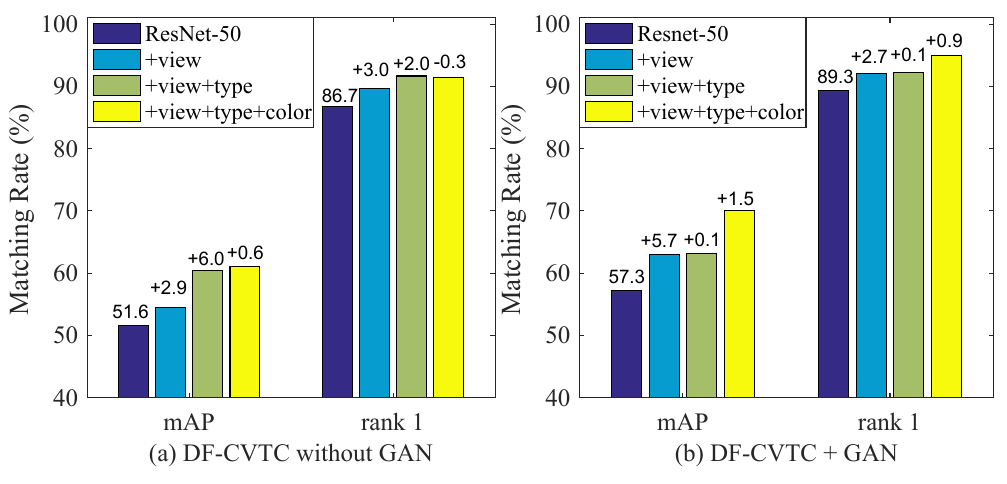}
 \caption{Performance of GAN on VeRi-776 dataset. (a) and (b) demonstrate the mAP and rank 1 scores of the proposed DF-CVTC and its variants without and with GAN respectively. The digits on the top of the last three bars on each metric indicate the degree of improvement by progressively introducing view, type and color, comparing to the first blue bar of the baseline ResNet-50. }\label{fig::vsgan}
\end{figure}

\begin{table*}[]
\caption{Ablation study on different backbones with varying components on VeRi-776 dataset  (in \%). The top three results are highlighted in {\color{red} red}, {\color{green} green} and {\color{blue} blue}, respectively.}\label{tb::otherbackbone}
\begin{center}
\begin{tabular}{|c|c|l|l|l|l|l|l|l|l|l|l|l|}
\hline
\multicolumn{1}{|r|}{\bf Component} &\multicolumn{3}{c|}{(a)} &\multicolumn{3}{c|}{(b)} &\multicolumn{3}{c|}{(c)} &\multicolumn{3}{c|}{(d)}\\
\hline
\multicolumn{1}{|c|}{view} &\multicolumn{3}{c|}{$\times$} &\multicolumn{3}{c|}{\checkmark} &\multicolumn{3}{c|}{\checkmark} &\multicolumn{3}{c|}{\checkmark}\\
\multicolumn{1}{|c|}{type} &\multicolumn{3}{c|}{$\times$} &\multicolumn{3}{c|}{$\times$} &\multicolumn{3}{c|}{\checkmark} &\multicolumn{3}{c|}{\checkmark}\\
\multicolumn{1}{|c|}{color} &\multicolumn{3}{c|}{$\times$} &\multicolumn{3}{c|}{$\times$} &\multicolumn{3}{c|}{$\times$} &\multicolumn{3}{c|}{\checkmark}\\
\cline{1-13}
{\bf Backbone} & mAP &\multicolumn{1}{c|}{rank 1} &\multicolumn{1}{c|}{rank 5} &mAP &rank 1 &rank 5 &mAP &rank 1 &rank 5 &mAP &rank 1 &rank 5\\
\hline
VGG16~\cite{Simonyan15} &\multicolumn{1}{l|}{\color{blue}42.35} &\color{blue}77.77 &\color{blue}88.14 &\color{blue}44.17 &\color{blue}80.63 &\color{blue}89.57 &\color{blue}45.43 &\color{blue}81.17 &\color{blue}90.35 &\color{blue}45.62 &\color{blue}81.76 &\color{blue}91.12\\
\hline
MobileNet~\cite{howard2017mobilenets} &\multicolumn{1}{l|}{\color{red}52.55} &\color{red}86.23 &\color{red}94.10 &\color{red}54.48 &\color{green}87.60 &\color{red}93.92 &\color{green}58.49 &\color{green}89.15 &\color{green}94.64 &\color{green}59.23 &\color{green}89.45 &\color{green}94.87\\
\hline
Inception-v4~\cite{szegedy2017inception} &\multicolumn{1}{l|}{\color{green}49.78} &\color{green}84.62 &\color{green}91.90 &\color{green}52.74 &\color{red}87.66 &\color{green}93.68 &\color{red}59.49 &\color{red}89.27 &\color{red}94.76 &\color{red}60.50 &\color{red}89.51 &\color{red}95.47\\
\hline
\end{tabular}
\end{center}
\end{table*}

\subsubsection{Analysis on subnetworks}
Table~\ref{tb::component} reports the effective of a different component on VehicleID and VeRi-776 dataset.
Obviously, by introducing the view, type and color subnetworks progressively, the performance of our method is consistently improved.
Compare to the base model of ResNet-50, our DF-CVTC make great progress in all metric.
In specific,  we increase the rank1 by 7.48\%, 6.36\% and 7.01\% in three difference scaler test sets respectively in VehicleID.
And increase the rank 1 by 4.65\%, mAP by 9.46\% respectively in VeRi-776.

Fig.~\ref{fig::progress} demonstrates an example of ranking results of the proposed DF-CVTC for a query from VeRi-776 dataset by progressively introducing the view, type and color subnetworks into the ResNet-50 backbone.
We observe that: 1) By introducing the view subnetwork, it can eliminate the wrong ranks with quite similar visible appearance especially with similar views to the query especially, such as rank 1 and rank 2 in Fig.~\ref{fig::progress} (c1).
2) By further introducing the type subnetwork, it can eliminate the wrong ranks with obviously distinct types, such as rank 2 and rank 9 in Fig.~\ref{fig::progress} (c2).
3) Our full model DF-CVTC (Fig.~\ref{fig::progress} (c4)) hits the rightest ranks by progressively introduce the three subnetworks.

Fig.~\ref{fig::progress} further visualizes the feature maps via CAM (Class Activation Mapping)~\cite{zhou2016learning} to demonstrate the attended area by progressively introducing the attribute subnetworks.
From which we can see, by introducing the view subnetwork, our method tends to emphasize the common area with different views from the query, such as the right hit rank 1 in as shown in Fig.~\ref{fig::progress} (c2).
Furthermore, progressively introducing the type subnetwork leads to higher attention on the areas reflecting the vehicle type information, comparing Fig.~\ref{fig::progress} (c3) to Fig.~\ref{fig::progress} (c2).
Similarly, introducing the color subnetwork leads to higher attention on discriminative color regions, as shown in Fig.~\ref{fig::progress} (c4).

\subsubsection{Analysis on backbones}
As we mentioned in Section~\ref{sec:backbone}, any other CNN architecture could be used in our framework instead of ResNet-50 without any limitation.
We further evaluate three prevalent CNN architectures, Inception-v4 \cite{szegedy2017inception}, VGG16 \cite{Simonyan15} and MobileNet \cite{howard2017mobilenets} as the backbone respectively while remaining the other part of the proposed model unchanged.
The results on VeRi-776 dataset are reported in Table~\ref{tb::otherbackbone}. From which we can see, all the three CNN counterparts achieve  satisfactory performance.
Specifically, Inception-v4 and MobileNet achieve competitive performance on all the metrics.
VGG16 works slight worse than the other two architecture, but it is still competitive to the state-of-the-art methods, which demonstrates that the high performance of the proposed model is not totally due to the superiority of the ResNet-50.
Furthermore, by progressively introducing the view, type and color subnetworks, the performance of the corresponding variants based on all the backbones consistently improves, which verifies the contribution of the proposed jointly learning model.

\subsubsection{Analysis on Attribute Predictors}
Table~\ref{tb::fixed} reports the result of our method while fixing the weighs of three attributes.
Specifically, we fix the weights of the view, type and color subnetworks as 1/5, 1/9 and 1/10, respectively.
After progressively training the model, we can find that with the introduction of the three subnetworks, there is no significant improvement.
The proposed DF-CVTC is even inferior than the baseline, which implies the importance of the adaptive weights learning during the proposed method.

\begin{table}[t]
\centering
\caption{Evaluation on our method while fixing the weights of three attributes.}\label{tb::fixed}
\setlength{\tabcolsep}{1.6mm}{
\begin{tabular}{l|ccc}
\hline
&\multicolumn{3}{c}{\bf VeRi-776}\\
\hline
Method &mAP & rank 1 & rank 5\\
\hline
ResNet-50 (baseline)  &51.58 &86.71 &92.43    \\
~~+view    &51.13	&87.66	&94.82\\
~~+view+type &52.02	&86.47	&93.50 \\
~~+view+type+color (DF-CVTC)  &50.09	&84.33	&92.31\\
\hline
\end{tabular}}
\end{table}

\begin{figure}[]
\centering
\includegraphics[width =9cm]{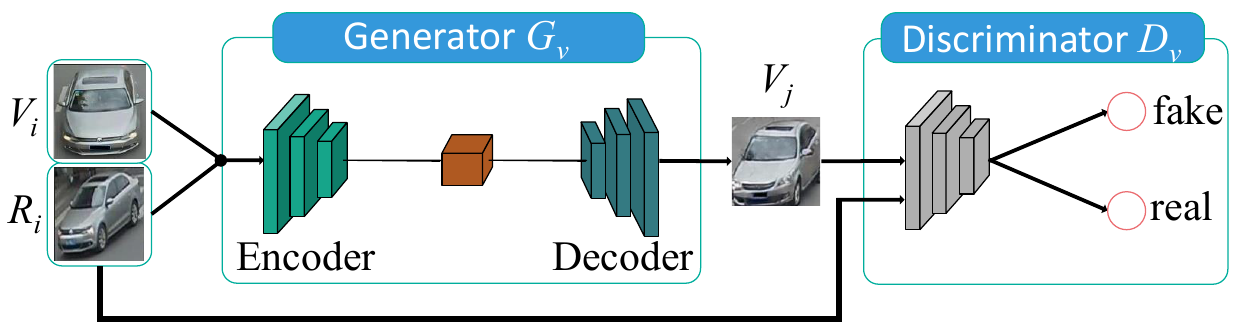}
\caption{The architecture of GAN based on the architecture of pix2pix~\cite{isola2017image}. For the input single view vehicle image {\bf $V_i$} ($front$ view as shown), it aims to synthesize a vehicle image {\bf $V_j$} with the same view as the target vehicle image {\bf $R_i$} ($front\_side$ view as shown). }\label{fig::VSGAN}
\end{figure}

\subsection{Data Augmentation}
\begin{figure}[t]
\centering
\includegraphics[width=9cm]{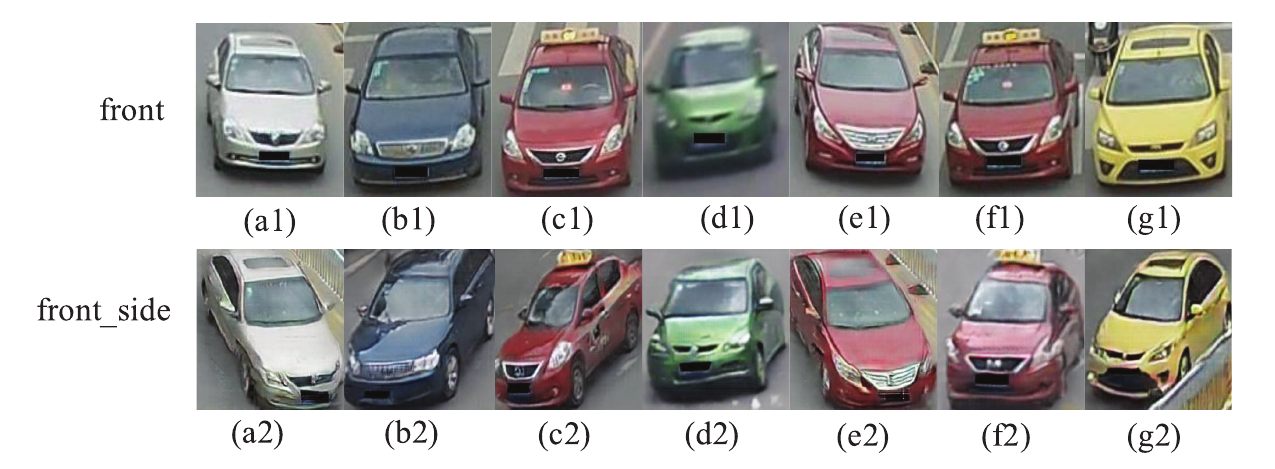}
 \caption{Examples of synthesizing the $front$ view vehicle images to $front\_side$ view on VeRi-776 dataset via GAN. The first and the second rows indicate the vehicle images with the original $front$ view and the synthesized $front\_side$ view respectively.}\label{fig::demogan}
\end{figure}

As we observed, most of the vehicle images under a certain camera are almost in the same viewpoint due to the rigid motion of vehicles, and thus the number of vehicle images with different views is very limited which brings a big challenge to train deep networks.
To handle this issue, we design a generative adversarial network (GAN) to generate the multi-view vehicle images.
In this paper, we simply employ pix2pix [45] for its generality. The generation architecture is illustrated as Fig.~\ref{fig::VSGAN}. In a specific, given an input vehicle image {\bf $V_i$} and a target vehicle image {\bf $R_i$} with a slightly different view the same ID, our GAN aims to generate a new vehicle image {\bf $V_j$} with the same view as {\bf $R_i$}. GAN constitutes a Generator {\bf $G_v$} learning a map conditional on the given target, and a Discriminator {\bf $D_v$} discriminating real data samples from the generated samples, such that the distribution of image {\bf $V_j$} is indistinguishable from the distribution image {\bf $V_i$}. The loss function can be expressed as,
\begin{equation}
\begin{split}
\mathcal{L}(G_v,D_v)=\mathbb{E}_{V_i,R_i}[logD_v(V_i,R_i)]+\mathbb{E}_{V_i,R_i}[log(1-\\
~D_v(V_i,G_v(V_i,R_i))]+\lambda\mathbb{E}_{V_i,R_i}[\Vert{R_i-G_v(V_i,R_i)}\Vert_1]\\
\end{split}
\end{equation}

\noindent where $G_v$ tries to minimize this objective against an adversarial $D_v$ that tries to maximize it, $\ell_1$ distance is used to encourage less blurring. $\lambda$ is the weighting coefficient. Fig.~\ref{fig::demogan} demonstrates several examples of synthesizing the $front$ view vehicle images to $front\_side$ view on VeRi-776 dataset via GAN. One more thing we would like to mention is the pair of the input images of our pix2pix GAN is from the same ID but slightly different viewpoint.

Due to the computational complexity, we have simply transferred 1400 front view vehicles into the front side view images for training data augmentation on the VeRi-776 dataset as shown in Fig.~\ref{fig::demogan}. %
Fig.~\ref{fig::vsgan} demonstrates the performance of GAN. From which we can see, by augmenting even only 1400 synthetic multi-view images into a total of 37729 training samples, it can benefit the Re-ID model with various components.
Moreover, it can further boost the contribution of the view subnetwork by improving $5.7\%$ and $2.7\%$ in mAP and rank 1 respectively, comparing to $2.9\%$ and  $3.0\%$ improvements of DF-CVTC without GAN.
We believe that more generated images with more viewpoints will further boost the performance.


\section{Conclusion}
In this paper, we have proposed a novel end-to-end deep convolutional network to jointly learn deep features, camera views, types and colors for vehicle Re-ID.
We expand the backbone of ResNet-50 with three consolidated subnetworks incorporating the view, type and color cues respectively.
These three tasks benefit each other and learn an informative discriminative representation for vehicle Re-ID.
Furthermore, we have increased the diversity of the views for vehicle images via a generative adversarial network.
By jointly learning the deep features, camera views, vehicle types and vehicle colors in a single unified framework, our method can achieve superior performance comparing to the state-of-the-art methods.
Comprehensive evaluation on two benchmark datasets demonstrates the clear contribution of each subnetwork and the capability of informative representation for vehicle Re-ID.


\ifCLASSOPTIONcaptionsoff
  \newpage
\fi



\bibliographystyle{IEEEtran}
%
%
%
%
\bibliography{vehicle}

%






\end{document}